\documentclass[final]{cvpr}

\usepackage{times}
\usepackage{epsfig}
\usepackage{graphicx}
\usepackage{amsmath}
\usepackage{amssymb}
\usepackage[utf8]{inputenc} 
\usepackage[T1]{fontenc}    
\usepackage{url}            
\usepackage{booktabs}       
\usepackage{amsfonts}       
\usepackage{nicefrac}       
\usepackage{microtype}      
\usepackage{graphicx}
\usepackage{multirow}
\usepackage{wrapfig}
\usepackage{xcolor}
\usepackage{bbm}
\usepackage{amsmath}
\usepackage{comment}
\usepackage{bbm}
\usepackage{footnote}
\newcommand\blfootnote[1]{%
  \begingroup
  \renewcommand\thefootnote{}\footnote{#1}%
  \addtocounter{footnote}{-1}%
  \endgroup
}

\usepackage[pagebackref=true,breaklinks=true,colorlinks,bookmarks=false]{hyperref}

\def\cvprPaperID{6073} 
\def\confYear{CVPR 2021}

\begin{document}

\title{3D-to-2D Distillation for Indoor Scene Parsing}

\author{Zhengzhe Liu$^{1}$ \quad   Xiaojuan Qi$^{2*}$ \quad     Chi-Wing Fu$^{1*}$ \\
$^1$The Chinese University of Hong Kong \quad $^2$The University of Hong Kong\\
{\tt\small \{zzliu,cwfu\}@cse.cuhk.edu.hk \quad  xjqi@eee.hku.edu.hk}
}

\maketitle

\begin{abstract}

\blfootnote{*: Corresponding authors}

Indoor scene semantic parsing from RGB images is very challenging due to occlusions, object distortion, and viewpoint variations. 
Going beyond prior works that leverage geometry information, typically paired depth maps, we present a new approach, a 3D-to-2D distillation framework, that enables us to leverage 3D features extracted from large-scale 3D data repositories (\eg, ScanNet-v2) to enhance 2D features extracted from RGB images.
Our work has three novel contributions.
First, we distill 3D knowledge from a pretrained 3D network to supervise a 2D network to learn  simulated 3D features from 2D features during the training, so the 2D network can infer without requiring 3D data.
%
Second, we design a two-stage dimension normalization scheme to calibrate the 2D and 3D features for better integration.
%
Third, we design a semantic-aware adversarial training model to extend our framework for training with unpaired 3D data.
%
Extensive experiments on various datasets, ScanNet-V2, S3DIS, and NYU-v2, demonstrate the superiority of our approach.
Also, experimental results show that our 3D-to-2D distillation improves the model generalization.

\end{abstract}

\section{Introduction}

\begin{figure}[t]
\centering
\includegraphics[width=0.99\columnwidth]{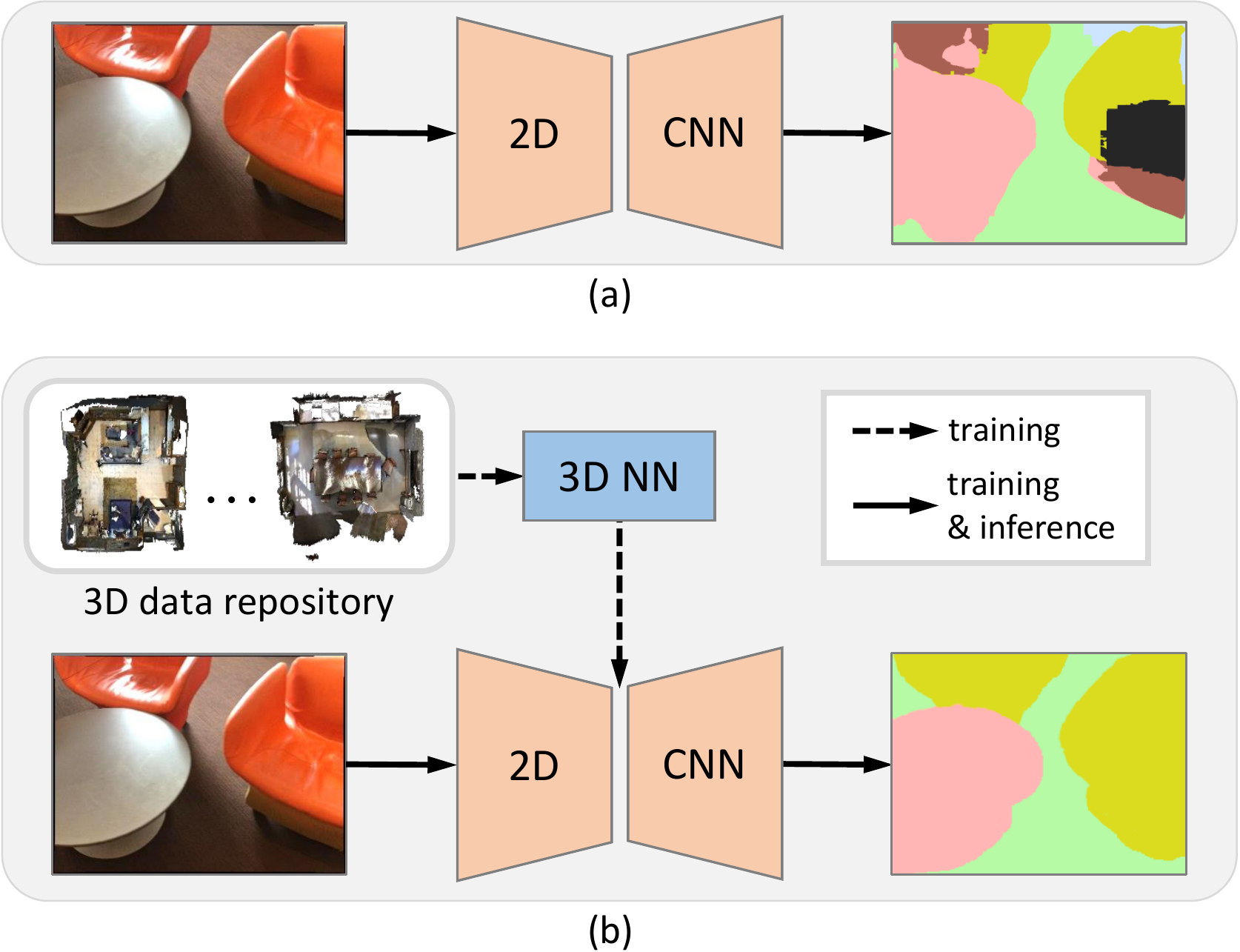}
\caption{
Compared with extracting features solely from the input image (a) for semantic parsing, our new approach (b) efficiently distills 3D features learned from a large-scale 3D data repository to train the 2D CNN to learn to enhance its features for better semantic parsing.
Our framework needs point cloud inputs only in training but not in testing, and the point cloud can be paired or unpaired.
}
\label{fig:overview_simple}
\end{figure}

Indoor scene parsing from images plays an important role in many applications such as robot navigation and augmented reality.
Though a considerable amount of advancements have been obtained with convolutional neural networks, this task is still very challenging, since the task inherently suffers from various issues, including distorted object shapes, severe occlusions, viewpoint variations, and scale ambiguities.

One approach to address the issues is to leverage auxiliary geometric information to obtain structured information that complements the RGB input.
For the auxiliary input, existing methods typically employ the depth map that associates with the input RGB image.
However, earlier methods~\cite{gupta2014learning,eigen2015predicting,cheng2017locality,qi20173d,romera2017erfnet,valada2019self,deng2019rfbnet} require the availability of the depth map inputs not only in the training but also in the testing.
As a result, they have limited applicability to general situations, in which depth is not available.
This is in contrast to the ubiquity of 2D images, which can be readily obtained by the many photo-taking devices around us.
%

To get rid of the constraint, 
several methods~\cite{wang2015towards,zhang2018joint,xu2018pad,zhang2019pattern,jiao2019geometry} propose to predict a depth map from the RGB input, then leverage the predicted depth to boost the scene parsing performance.
However, depth prediction from a single image is already a very challenging task on its own.
Hence, the performance of these methods largely depends on the quality of the predicted depth.
Also, the additional depth prediction raises the overall complexity of the network.


Besides the above issues, a common limitation of the prior works is that they only explore the depth map as the auxiliary geometry cue.
Yet, a depth map can only give a partial view of a 3D scene, so issues like occlusions and viewpoint variations are severe.
Further, they all require paired RGB-depth data in training.
So, they are limited for use on datasets with depth maps, which require tedious manual preparation,~\eg, hardware setups,  complicated calibrations, etc.

In this work, we present the first flexible and lightweight framework (see Figure~\ref{fig:overview_simple}), namely \textit{3D-to-2D distillation}, to distill occlusion-free, viewpoint-invariant 3D representations derived from 3D point clouds for embedding into 2D CNN features by training the network to learn to simulate 3D features from the input image.
Our approach leverages existing large-scale 3D data repositories such as ScanNet-v2~\cite{dai2017scannet} and S3DIS~\cite{armeni2017joint} and recent advancements in 3D scene understanding~\cite{graham2015sparse,choy20194d,han2020occuseg,jiang2020pointgroup} for 3D feature extraction, and allows the use of unpaired 3D data to train the network.
%

%
%

For the 2D CNN to effectively learn to simulate 3D features, our 3D-to-2D distillation framework incorporates a two-stage \textit{dimension normalization} (DN) module to explicitly align the statistical distributions of the 2D and 3D features.
So, we can effectively reduce the numerical distribution gap between the 2D and 3D features, as they are from different data modalities and neural network models. 
Also, a \textit{Semantic Aware Adversarial Loss} (SAAL) is designed to serve as the objective of model optimization without paired 2D-3D data to make the framework flexible to leverage existing 3D data repository and boost its applicability.

We conduct extensive experiments on  indoor scene parsing datasets ScanNet-v2~\cite{dai2017scannet}, S3DIS~\cite{armeni2017joint}, and NYU-v2~\cite{silberman2012indoor}. With only a negligible amount of extra computation cost, 
 our approach consistently outperforms the baselines including the state-of-the-art depth-assisted semantic parsing approach~\cite{jiao2019geometry} and our two baselines that leverage depth maps, manifesting the superiority of our approach.
Besides, our further in-depth experiments on a depth reconstruction task implies that our framework can effectively embed 3D representations into 2D features and produce much better reconstruction results.
More importantly, our model obtains a significant performance gain (19.08\% {\vs} 27.22\% mIoU), even when evaluated on data from an unseen domain, suggesting that the 3D information embedded by our 3D-to-2D distillation helps promote the generalizability of CNNs.


\begin{figure*}
\centering
\includegraphics[width=1\textwidth]{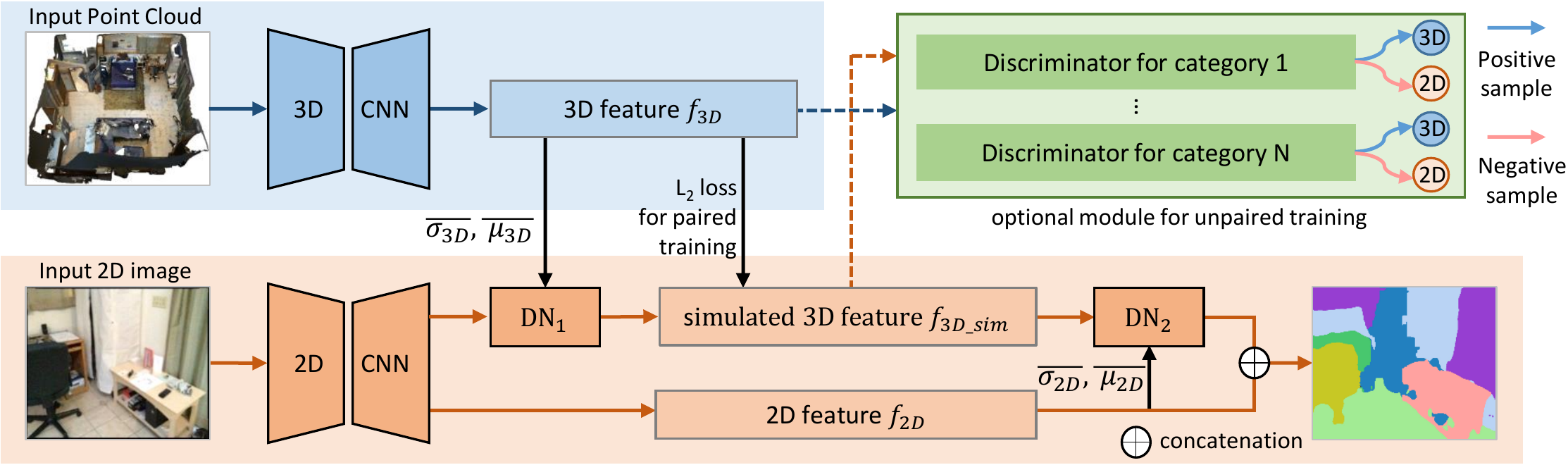}
\caption{Overview of our 3D-to-2D distillation framework for 2D semantic segmentation.
During the training, the framework takes a 2D image and a 3D point cloud as inputs.
We transfer 3D feature $f_{\text{3D}}$ from the 3D network (blue) to the 2D network (orange) with an $L_2$ loss (for paired 2D-3D data), such that the 2D network can learn to produce simulated 3D feature $f_{\text{3D}\_\text{sim}}$ from the 2D image and we do not need $f_{\text{3D}}$ and 3D point cloud input during the inference.
Also, note the two dimension normalization modules ($\text{DN}_1$ and $\text{DN}_2$) for aligning 2D-3D features, and the optional module on top-right for training with unpaired 2D-3D data using the semantic-aware adversarial loss.}
\label{fig:overview}
\vspace*{-1.5mm}
\end{figure*}


\section{Related Work}


\paragraph{Semantic segmentation.}
The computer vision community has gained remarkable achievements on semantic segmentation~\cite{long2015fully,yu2015multi,paszke2016enet,qi2016augmented,badrinarayanan2017segnet,chen2017deeplab,zhao2018icnet,zhao2017pyramid,wang2018non,zhao2018psanet,yuan2018ocnet,yuan2019ocr,wang2020deep}.
PSPNet~\cite{zhao2017pyramid} is a representative work that has inspired many follow-ups.
In this work, we adopt PSPNet as the baseline model for semantic segmentation, as it has an open-source repository with good reproducibility and delivers competitive performance even compared with the latest works.

\vspace*{-3.5mm}
\paragraph{3D semantic segmentation.}
%
%
Methods for 3D semantic segmentation are generally point-based or voxel-based.
\textit{Point-based networks} adopt raw point clouds as input.
Along this line of works,~\cite{qi2017pointnet,qi2017pointnetv2} are pioneering ones. 
Later, various convolution-based methods~\cite{li2018pointcnn,thomas2019kpconv,wu2019pointconv,boulch2020convpoint} were proposed for 3D semantic segmentation on point clouds. Recently, Kundu~\etal~\cite{kundu2020virtual} proposed to fuse features from multiple 2D views for 3D semantic segmentation.

On the other hand, \textit{Voxel-based networks} first voxelize the raw data into regular 3D grids for feature learning~\cite{riegler2017octnet,graham2015sparse,su2018splatnet,dai20183dmv,zhao2018modeling}.
The recently-proposed methods MinkowskiNet~\cite{choy20194d} and OccuSeg~\cite{han2020occuseg} are two of the representative works in this branch.   
In this work, we adopt PointWeb~\cite{zhao2019pointweb} and MinkowskiNet~\cite{choy20194d} as the architectures for extracting point-based and voxel-based 3D features, respectively. 

\vspace*{-3.5mm}
\paragraph{Knowledge distillation.}
Our work shares a similar spirit as knowledge distillation techniques~\cite{hinton2015distilling,romero2014fitnets,gupta2016cross,mirzadeh2019improved,garcia2018modality,heo2019knowledge,liu2020structured} in that we both aim to transfer features from a source model (i.e., the teacher model in knowledge distillation or the 3D network in our work) to a target model (\ie, the student model in knowledge distillation or the 2D network in our work) to enhance the performance of the target model. However, conventional knowledge distillation techniques are designed typically for scenarios, in which
(i) the source data to be distilled has the same modality~\cite{hinton2015distilling,romero2014fitnets,gupta2016cross,mirzadeh2019improved} or similar modalities~\cite{garcia2018modality} as the target,
(ii) the two networks for the feature extraction have the same or similar architecture (\eg, convolution neural networks),
and
(iii) the distillation objective typically requires paired source-target data.

Recently, cross-modality distillation has also been studied in~\cite{pande2019adversarial,garcia2019learning,roheda2018cross}. However, their methods are not suitable in our, since they cannot distinguish features of different categories. To this end, we propose to associate 2D and 3D features by the object category and formulate the SAAL to enable unpaired training. 
The experimental result shows that our approach outperforms the most recent semantic segmentation knowledge distillation method~\cite{liu2020structured}.


\section{Method}

This section presents our proposed 3D-to-2D distillation framework for effective distillation of 3D features learned from point clouds to improve the performance of 2D indoor scene parsing.
Figure~\ref{fig:overview} shows the overall architecture of the framework.
During the training, our framework takes a 2D image and a 3D point cloud as its inputs.
To begin, we use a 3D CNN to extract 3D feature $f_\text{3D}$ from the input 3D point cloud (blue region in Figure~\ref{fig:overview}).
On the other hand, we use a 2D CNN to extract 2D feature $f_\text{2D}$ from the input image, and produce also simulated 3D feature $f_{\text{3D}\_\text{sim}}$ in the 2D network (orange region in Figure~\ref{fig:overview}).
For the 2D CNN, we adopt PSPNet~\cite{zhao2017pyramid} for the case of semantic segmentation, whereas for the 3D CNN, we may adopt different 3D architectures, such as PointWeb~\cite{zhao2019pointweb} and MinkowskiNet~\cite{choy20194d}.
Also, note that $\text{DN}_1$ and $\text{DN}_2$ are our dimension normalization modules.

An important insight in our approach is that during the training, we use $f_\text{3D}$ to supervise the generation of $f_{\text{3D}\_\text{sim}}$, so that the 2D network can learn to produce $f_{\text{3D}\_\text{sim}}$ that looks like $f_{\text{3D}}$.
In this way, we do not need the point cloud input and 3D network when we use our framework to test on a 2D image; the 2D network alone can generate $f_{\text{3D}\_\text{sim}}$ solely from the image input.
Also, to resolve the statistical difference between the 2D and 3D inputs, we design a two-stage dimension normalization module ($\text{DN}_1$ and $\text{DN}_2$) to effectively transform features before and after $f_{\text{3D}\_\text{sim}}$ (Figure~\ref{fig:overview}), so that the statistical distributions of the 2D and 3D features in the 2D network are better aligned to facilitate the learning of the simulated 3D feature and also its integration with the remaining part of the 2D network.
Further, we design a semantic-aware adversarial loss as an optional module in the training (Figure~\ref{fig:overview} (top right )) to extend our framework for training with 2D-3D inputs that are unpaired.

In this section, we first present how our framework is trained with paired 2D-3D inputs (Section \ref{sec:pair}).
Then, we present our dimension normalization modules in Section~\ref{sec:normalization} and how we extend the framework for unpaired training with the semantic aware adversarial loss in Section~\ref{sec:adverse}.



\subsection{Training Objectives with Paired 2D-3D Data}\label{sec:pair}


Since 3D feature $f_{\text{3D}}$ extracted from the input point cloud is point-wise in the 3D space, we cannot directly associate it with 2D feature $f_{\text{2D}}$ and likewise the simulated 3D feature $f_{\text{3D}\_\text{sim}}$, which are both defined in 2D image space.
To determine the associations, if paired 2D-3D data is available in the training, we can use the given camera parameters to transform and project the 3D points to the 2D image space.
In this way, we can determine the pixel location associated with each point in the input point cloud, and obtain  $f_{\text{3D}\_{i}}$, which denotes the projected 3D feature at pixel $i$.
If more than one point projects to the same pixel, we consider only the point that is the nearest to the camera, since it should correspond to the visible pixel in the image input.

Then, we adopt an $L_2$ regression loss between $f_{\text{3D}}$ and $f_{\text{3D}\_\text{sim}}$ to supervise the generation of $f_{\text{3D}\_\text{sim}}$ in the 2D network.
Clearly, the projected 3D points are sparse in the image space, so we locate the pixels covered by the points and perform the regression only on the covered pixels:
\begin{equation} 
L_p=\sum_{i} ||f_{\text{3D}\_{i}}- f_{\text{3D}\_\text{sim}\_{i}}||^2_2,
\label{equ:l2}
\end{equation}
where $i$ indexes the pixels covered by the projected 3D points and $L_p$ denotes the loss to guide the generation of $f_{\text{3D}\_\text{sim}}$.

To train the whole framework for the semantic segmentation task, we employ the cross-entropy loss below:
\begin{equation}
L_s=\sum_{i}\sum_{c} -\mathbbm{1}_{i,c} \log{p_{i,c}},
\label{equ:seg}
\end{equation}
where
$p_{i,c}$ denotes the probability of pixel $i$ belonging to category $c$ 
and
$\mathbbm{1}_{i,c}$ is an indicator function that equals $1$ if the ground-truth category of pixel $i$ is  $c$; otherwise, it is zero.


\subsection{Dimension Normalization}\label{sec:normalization}

\begin{table*}
\centering
\scalebox{0.8}{
  \begin{tabular}{c|c|c}
    \toprule
     BN & AdaBN & DN\\
    \midrule
     \large{ $\hat{x}=\gamma \frac{x-\mu_\text{2D} }{\sigma_\text{2D}} +\beta$ } & \large{$\hat{x}=\gamma \frac{x-\bar \mu_\text{3D} }{\bar\sigma_\text{3D}} +\beta$} & \large{$\hat{x}_\text{3D}=\Delta \sigma_\text{3D}(\bar\sigma_\text{3D} \frac{x-\mu_\text{2D} }{\sigma_\text{2D}} +\bar\mu_{3D}) +\Delta \mu_{3D} $}, \large{$\hat{x}_\text{2D}=\Delta \sigma_\text{2D}(\bar\sigma_\text{2D} \frac{x-\mu_\text{3D} }{\sigma_\text{3D}} +\bar\mu_\text{2D}) +\Delta \mu_\text{2D} $}\\ 
    \bottomrule
  \end{tabular}
}
\vspace{0.05in}\caption{Batch Normalization (BN), Adaptive Batch Normalization (AdaBN), and Dimension Normalization (DN). $\gamma$, $\beta$ and $\Delta$ are learnable parameters, whereas $\mu$ and $\sigma$ indicate mean and variance.} 
\label{tab:dndef}
\vspace*{-1.5mm}
\end{table*}

To resolve the statistical difference between 2D and 3D representations induced by different data modalities and neural network architectures, we design the dimension normalization modules $\text{DN}_1$ and $\text{DN}_2$ to explicitly calibrate the distribution of the 2D and 3D features; see Figure~\ref{fig:overview}.

Figure~\ref{fig:dn} illustrates the technical details.
Give a batch of $N$ feature maps as inputs to $\text{DN}_1$ or $\text{DN}_2$, each of the dimensions $H$ (height), $W$ (width), and $C$ (channels), we compute channel-wise means and variances of the feature map over the $N$, $H$, and $W$ dimensions: ($\mu_\text{2D}, \sigma_\text{2D}$) for the input 2D feature map to $\text{DN}_1$ or ($\mu_\text{3D}, \sigma_\text{3D}$) for the input 3D feature map to $\text{DN}_2$.
On the other hand, as shown in Figure~\ref{fig:overview},
$\text{DN}_1$ receives statistics ($\bar\mu_\text{3D}$, $\bar\sigma_\text{3D}$) of the 3D feature $f_\text{3D}$, whereas $\text{DN}_2$ receives statistics ($\bar\mu_\text{2D}$, $\bar\sigma_\text{2D}$) of the 2D feature $f_\text{2D}$.
Below, we detail the procedure inside each DN:
%
%
\begin{itemize}
\item
The purpose of $\text{DN}_1$ is to transform the input 2D feature to align its distribution with that of the 3D feature for effective feature learning.
Here, we first use two convolution layers of $3$$\times$$3$ and $1$$\times$$1$ ($\text{Conv}_1$ in Figure~\ref{fig:dn}) to transform the input features to have the same channel dimension as the penultimate layer in 3D network $f_\text{3D}$.
Then, we normalize the transformed features by $\mu_\text{2D}$ and $\sigma_\text{2D}$, and use $\bar\mu_\text{3D}$ and $\bar\sigma_\text{3D}$ to scale and adjust the normalized features.
Ideally, $\bar\mu_\text{3D}$ and $\bar\sigma_\text{3D}$ should correspond to $f_\text{3D}$.
However, $f_\text{3D}$ is available only in the training but not in the inference, so we pre-compute $\bar\mu_\text{3D}$ and $\bar\sigma_\text{3D}$ globally over the entire data, and use the pre-computed values in both training and inference.
As a result, we further use learnable parameters $\Delta\sigma_\text{3D}$ and $\Delta\mu_\text{3D}$ in $\text{DN}_1$ to adjust the features, followed by two subsequent convolution layers of $3$$\times$$3$ and $1$$\times$$1$ ($\text{Conv}_2$ in Figure~\ref{fig:dn}) to produce the simulated 3D features.

\begin{figure}
\centering
\includegraphics[width=0.95\columnwidth]{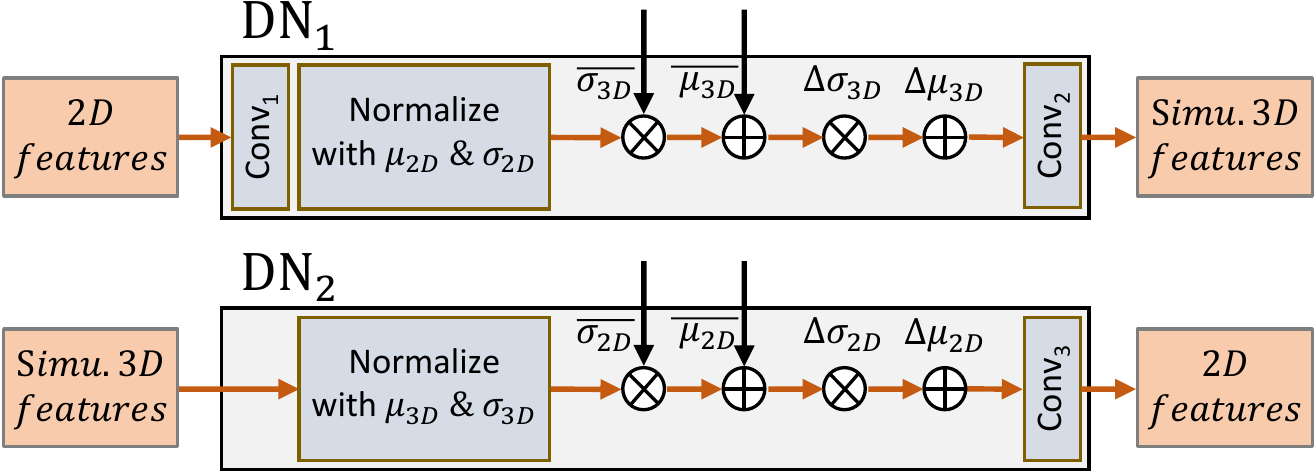}  
\caption{Dimension normalization modules.
Note that $\otimes$ means element-wise multiplication;
$\oplus$ means element-wise addition; and
$\text{Conv}_1$ to $\text{Conv}_3$ are additional convolutional layers.
} \label{fig:dn}
\vspace*{-2.5mm}
\end{figure}

%
\vspace*{-1.5mm}
\item
The purpose of $\text{DN}_2$ is to calibrate the learned 3D feature $f_\text{3D\_sim}$ back to $f_\text{2D}$ for smooth 2D-3D feature concatenation.
Its structure follows that of $\text{DN}_1$ (Figure~\ref{fig:dn}), except that we do not need any additional convolutional layer to pre-transform the input feature.
\end{itemize}

Different from standard batch normalization (BN)~\cite{ioffe2015batch}, which further learns a linear transform to enhance the representative capability as shown in Table~\ref{tab:dndef}, our 3D-to-2D distillation module modulates the normalized distribution with a global pre-calculated 3D statistics $\bar\sigma_\text{3D}$ and $\bar\mu_\text{3D}$ and then uses learnable offsets $\Delta\sigma_\text{3D}$ and $\Delta\mu_\text{3D}$ to further adjust the result.
With 3D-to-2D distillation, we can explicitly align the distributions of the 2D and 3D features.
Yet, our approach still retains the advantage of BN by normalizing features in a batch-wise manner during the training. Then, during the inference, $\mu_\text{2D}$ and $\sigma_\text{2D}$ are replaced by the accumulated $\bar \mu_\text{2D}$ and $\bar \sigma_\text{2D}$ similar to BN.


\vspace*{-3.5mm}
\paragraph{Relation to BN and AdaBN.} 
Fundamentally, BN is proposed to facilitate the training of deep neural networks by normalizing the data distribution in a batch as shown in Table~\ref{tab:dndef}.
It helps to reduce the internal co-variant shift~\cite{ioffe2015batch} or smooth the objective function~\cite{santurkar2018does}.
However, BN is domain-dependent, which has side-effects in our cross-modality knowledge transfer task.
Further, AdaBN~\cite{li2016revisiting} is proposed for domain adaptation. In the inference stage, AdaBN uses ($\bar\mu_\text{3D}, \bar\sigma_\text{3D}$) of the target domain instead of the accumulated ($\bar\mu_\text{2D}, \bar\sigma_\text{2D}$) in the source domain to normalize the features. AdaBN is designed for the case, in which the model is trained in the source domain and tested in the target domain, which is different from our setup where training and inference are both performed in the 2D domain.



\subsection{Adversarial Training with Unpaired Data}\label{sec:adverse}
With paired 2D-3D data, we can distill 3D feature with an $L_2$ loss, since we can correspond the 2D and 3D features in the same image domain.
However, paired 2D-3D data are typically expensive to acquire in a large quantity, given the amount of works needed in data collection, calibration, and annotation.
To this end, we design a new adversarial training approach to correlate 2D and 3D features and supervise the generation of the simulated 3D features.

There are two key insights in our approach.
First, we observe that existing 2D and 3D datasets usually have common object categories, e.g., ScanNet-v2~\cite{dai2017scannet} has 20 object categories in its 3D point clouds, whereas NYU-v2~\cite{silberman2012indoor} has 40 object categories in its 2D images; the 20 categories in ScanNet-v2 can all be found in the categories in NYU-v2.
Hence, we propose to {\em correspond 2D and 3D features by their associated object categories\/}.
Second, given a 3D feature of a certain category and another 2D feature of the same category, we propose a novel adversarial training model with a per-category discriminator.
The goal of our model is to generate simulated 3D features solely from the 2D features, such that {\em the discriminator cannot differentiate the 3D features from the simulated one of the same category\/}.
In this way, the 2D network should learn to generate simulated 3D features solely without requiring paired 2D-3D data.

Figure~\ref{fig:overview} (top-right) shows the optional module we designed for unpaired training.
It has $N$ individual discriminators, where $N$ is the number of object categories common to the 2D and 3D data.
We implement each discriminator using six fully-connected layers and denote the discriminator of category $c$ as $D_c$.
There are two kinds of inputs to $D_c$: (i) a 3D feature vector $f_\text{3D\_i}$ from 3D network or (ii) a simulated 3D feature vector $f_\text{3D\_sim\_j}$ from 2D network, and both should belong to category $c$.
For each input to $D_c$, it should predict a confidence score that indicates whether the input feature vector comes from the 2D or 3D network.

Before we present how we train the whole framework for unpaired data, we first denote $\Phi_\text{2D}$ as the 2D network.
During the training, the 3D network is fixed, and we alternatively train $\Phi_\text{2D}$ and the set of discriminators $\{D_c\}$, similar to the way the generator (like $\Phi_\text{2D}$) and discriminator (like $D_c$'s) are trained in a conventional GAN model.
\begin{itemize}
\item
When we train the discriminators, we fix $\Phi_\text{2D}$ and use the following objective to train each $D_c$:
\begin{equation}
\begin{aligned}
L_\text{adv}(D_c) = - \sum_{i} \mathcal{N}_{c,i} \log (1-D_c(f_\text{3D\_sim\_i})) \\
- \sum_{j} \mathcal{M}_{c,j} \log (D_c(f_\text{3D\_j})),
\end{aligned}
\end{equation}
where $\mathcal{N}_{c,i}$ (or $\mathcal{M}_{c,j}$) equals $1$, if the 2D pixel $i$  (or 3D point $j$) belongs to category $c$; otherwise, it equals $0$.
Note that we treat $f_\text{3D\_sim\_i}$ as a negative sample and $f_\text{3D\_j}$ as a positive sample, so the goal of $D_c$ is to learn to differentiate them; by then, $\Phi_\text{2D}$ should learn to generate better $f_\text{3D\_sim\_i}$ to deceive $\{D_c\}$.
\item
When we train the 2D network $\Phi_\text{2D}$, we fix all $D_c$'s and use the following objective in the training:
\begin{equation}
\begin{aligned}
L_\text{adv}(\Phi_\text{2D}) = - \sum_{i} \sum_{c} \mathcal{N}_{c,i} \log (D_c(f_\text{3D\_sim\_i})).
\end{aligned}
\end{equation}
\end{itemize}
\vspace*{-2mm}
Overall, to train $\Phi_\text{2D}$, the overall loss is constructed by combining the softmax-cross entropy loss~\cite{zhao2017pyramid} for semantic segmentation with the paired regression loss $L_p$ or with the adversarial loss $L_\text{adv}(\Phi_\text{2D})$, if paired data is unavailable.
The loss weights are validated on a small validation set.

\section{Experiments and Results}\label{sec:experiments}
\paragraph{Datasets} We conduct experiments on three indoor scene parsing datasets---ScanNet-v2~\cite{dai2017scannet}, S3DIS~\cite{armeni2017joint}, and NYU-v2~\cite{Silberman:ECCV12}.  \textit{ScanNet-v2}~\cite{dai2017scannet} contains 1,513 scenes with 3D scans and 2D images.
\textit{S3DIS}~\cite{armeni2017joint} contains 3D scans with associated 2D images of 271 rooms, and each 3D point and 2D pixel are annotated as one of the 13 categories in the dataset.
To be noted, for both datasets, we can only obtain sparse paired 2D-3D points, due to the calibration and projection issues as illustrated in the supplementary material.
Only $16.38\%$ and $10.57\%$ of 2D pixels have corresponding 3D points in ScanNet-v2 and S3DIS, respectively.
\textit{NYU-v2}~\cite{Silberman:ECCV12} contains 1,449 images without reconstructed 3D scene.

\vspace*{-3.5mm}
\paragraph{Implementation details.} We implement our framework using PyTorch~\cite{NEURIPS2019_9015}, and train all the models and baselines with the SGD optimizer. The batch size is 16 and the initial learning rate is 0.01, which is scheduled based on the ``poly'' learning rate policy with power 0.9~\cite{zhao2017pyramid}.
Since different datasets have different number of training samples, we train models with different number of epochs on different datasets to equalize the number of training iterations:
ScanNet-v2 for 50 epochs;
S3DIS for 20 epochs on Area 1,2,3,4,6; and
NYU-v2 with the officially split training samples for 300 epochs.
By default, we empirically set 
the loss weights on $L_{\text{s}}$, $L_{\text{adv}}$ and
$L_p$ as 1, 0.01 and 0.03, respectively.

\vspace*{-3.5mm}
\paragraph{Network architectures.}
By default, we adopt the representative architecture PSPNet-50~\cite{zhao2017pyramid} for semantic segmentation. For 3D feature extraction, we employ the voxel-based architecture, {\ie}, MinkowskiNet~\cite{choy20194d} on ScanNet~\cite{dai2017scannet}, and the point-based architecture, {\ie}, PointWeb~\cite{zhao2019pointweb} on S3DIS. The 3D network weights are fixed during the training.



\begin{figure}[!t]
\centering
\includegraphics[width=0.95\columnwidth]{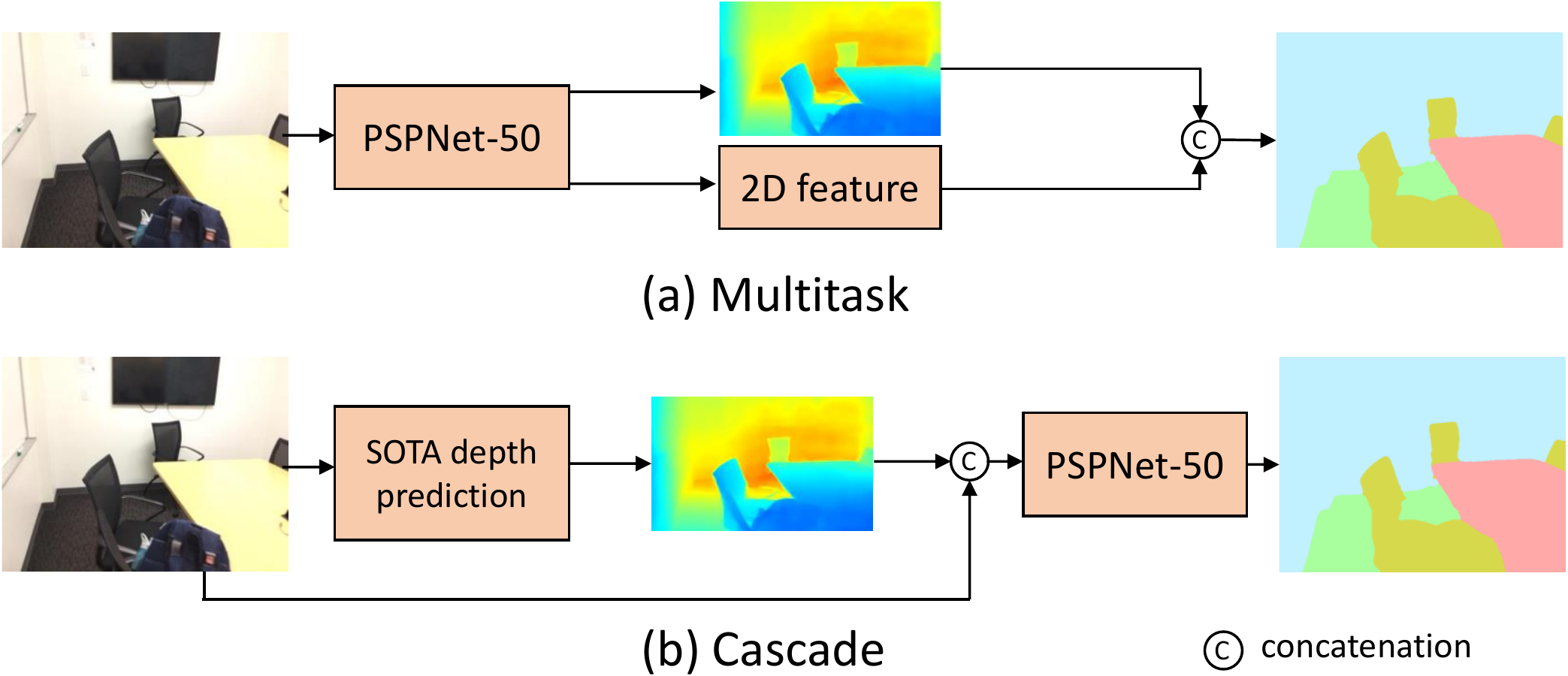}
\caption{Illustrating the Multitask and Cascade approaches of leveraging depth for 2D semantic segmentation.}
\label{fig:baseline}
\vspace*{-1.5mm}
\end{figure}

\subsection{Comparing with Related Geometry-Assisted and Knowledge Distillation Methods}

First, we present extensive experiments to compare our 3D-to-2D distillation approach with major competitors.
They are alternative ways of leveraging depth or 3D information to enrich 2D features for semantic segmentation.
In these experiments, we use the latest large-scale indoor scene parsing dataset ScanNet-v2~\cite{dai2017scannet}, which provides paired 2D-3D data for training various models for semantic segmentation.
%
%
\begin{itemize}
\vspace*{-1mm}
\item[(i)] Baseline: the original PSPNet-50 model~\cite{zhao2017pyramid}.
\vspace*{-1.75mm}
\item[(ii)] Multitask:
we use PSPNet-50 to additionally predict a depth map (Figure~\ref{fig:baseline} (a)), which is supervised by using data from ScanNet-v2; this model is similar to ours, in the sense that we replace the simulated 3D feature in our model by a depth map prediction.
%
\vspace*{-1.75mm}
\item[(iii)] Cascade: we use a state-of-the-art depth prediction method~\cite{lee2019big}\footnote{\url{https://paperswithcode.com/sota/monocular-depth-estimation-on-nyu-depth-v2}} to predict a depth map and take the predicted depth map as an auxiliary input to assist PSPNet (Figure~\ref{fig:baseline} (b)); so, this approach also does not require 3D data in the inference like ours, but the knowledge comes from the depth map predicted by~\cite{lee2019big}.
\vspace*{-1.75mm}
\item[(iv)] Geo-Aware~\cite{jiao2019geometry}: this is the most recent method that distills features from depth maps for assisting 2D semantic segmentation; since code is not available, we simply report its ScanNet-v2 result on its paper in Table~\ref{tab:scannet}.
Also, this method is finetuned on NYU-v2, so it leveraged more data than ours (ScanNet-v2 only).
%
\item[(v)] Structured KD: we adopt the most recent knowledge distillation approach for semantic segmentation~\cite{liu2020structured}; since it is designed for distilling 2D information, we replace its teacher net with a 3D network to distill 3D information for comparison with our approach.
\end{itemize}

Table~\ref{tab:scannet} reports the semantic segmentation performance of our approach and the competitors on the ScanNet-v2 validation set.
Our approach outperforms all of them, and these results reveal the following:
\begin{itemize}
\vspace*{-1.5mm}
\item 
Comparing with Multi-task, we can show that our full approach of predicting the simulated 3D features leads to better results than explicitly predicting a depth map using the same paired data.
This result demonstrates the richness of simulated 3D features from 3D point clouds, as compared with 2.5D depth maps.
\vspace*{-1.5mm}
\item
Comparing with Cascade, we can show that even we use the state-of-the-art depth prediction network~\cite{lee2019big} to provide the predicted depth map, the predicted depth map (which is 2.5D) cannot assist the PSPNet for semantic segmentation, as good as the simulated 3D features generated by our 3D-to-2D distillation approach.
\vspace*{-1.5mm}
\item
Comparing with Geo-Aware~\cite{jiao2019geometry}, we show again that distilling features from point clouds can lead to a better performance.
Note that Multitask, Cascade, and Geo-Aware are all depth-assisted methods.
\vspace*{-1.5mm}
\item
Comparing with Structured KD, we show that our 3D-to-2D distillation approach can lead to a higher performance than simply adopting a general knowledge distillation model.
To be noted, Structured KD also adopts an adversarial objective to align the global segmentation map; see Section~\ref{sec:ablation}.
%
\end{itemize}







\begin{table}
\centering
\scalebox{0.8}{
  \begin{tabular}{c|c|c}
    \toprule
     & Method & mIoU\\
     \midrule
     Baseline & PSPNet-50  & 53.40 \\
     \midrule
     & Multitask & 54.08 \\
     Depth methods & Cascade & 53.72 \\
     & Geo-Aware~\cite{jiao2019geometry}  & 56.90 \\
     \midrule
     KD & Structured KD &   56.36 \\
     \midrule
     Ours & 3D-2D Distillation & \textbf{58.22} \\
    \bottomrule
  \end{tabular}
}
\vspace*{0.05in}
\caption{Comparing the semantic segmentation performance (on ScanNet-v2 validation set) of various methods that use depth/3D information to assist the segmentation: two alternative methods (Multitask and Cascade) that predict depth maps, a depth-distillation method Geo-Aware~\cite{jiao2019geometry}, and Structured KD, which is a variant of~\cite{liu2020structured}. All the methods are based on the ResNet-50 backbone.} 
\label{tab:scannet}
\vspace*{-1mm}
\end{table}

\begin{table}
\centering
\scalebox{0.8}{
  \begin{tabular}{c|c}
    \toprule
    Method & mIoU\\
    \midrule
    FCN-8s~\cite{long2015fully} & 45.87\\
    ParseNet~\cite{liu2015parsenet} & 47.72\\
    DeepLab-v2~\cite{chen2017deeplab} & 43.89\\
    AdapNet~\cite{valada2017adapnet} & 47.28\\
    DeepLab-v3~\cite{chen2017rethinking} & 50.09\\
    AdapNet++~\cite{valada2019self} & 52.92\\
    Ours (w/ PSPNet-50) & \textbf{57.76}\\
    \bottomrule
  \end{tabular}
}
\vspace*{0.05in}
\caption{Comparing our method with existing 2D image-based semantic segmentation methods on the ScanNet-v2 validation set.
We follow the settings in~\cite{valada2019self}, where the size of the input image is 384$\times$768 without left-right flip and multi-scale test.}
\label{tab:2d}
\vspace*{-1mm}
\end{table}

\begin{table}
\centering
\scalebox{0.8}{
  \begin{tabular}{c|c}
    \toprule
     Method & mIoU\\
     \midrule
     HRNet~\cite{wang2020deep} + OCR~\cite{yuan2018ocnet}  & 60.56 \\
     \midrule
     HRNet~\cite{wang2020deep} + OCR~\cite{yuan2018ocnet} + Our 3D-2D distillation & \textbf{61.36} \\
    \bottomrule
  \end{tabular}
}
\vspace*{0.05in}
\caption{We adopt our method into the state-of-the-art 2D semantic segmentation approach, HRNet~\cite{wang2020deep} + OCR~\cite{yuan2018ocnet}, and further boost its performance.
The networks use HRNet-W48~\cite{wang2020deep} as the backbone, with a comparable network complexity as ResNet-101.}
\label{tab:hrnet}
\vspace*{-1mm}
\end{table}




\subsection{Further Evaluations with Sparse Paired Data}

Next, we conduct further experiments on the latest large-scale indoor scene parsing dataset---ScanNet-v2 and S3DIS.

\vspace*{-3.5mm}
\paragraph{ScanNet-v2 semantic segmentation.}~\label{sec:scannet} 
First, we compare our method with several 2D image-based semantic segmentation approaches.
Table~\ref{tab:2d} reports the results, showing that our model outperforms all of them for 4.8\% to 11.8\% mIoU.

Further, since our method is generic, we can easily incorporate it into existing semantic segmentation architecture.
Hence, we adopt it into the state-of-the-art 2D semantic segmentation network---HRNet+OCR~\cite{yuan2019ocr,wang2020deep}, which has demonstrated top performance on various datasets.
As shown in Table~\ref{tab:hrnet}, our 3D-to-2D distillation approach can further boost the performance of HRNet+OCR.

\begin{table}
\centering
\scalebox{0.8}{
  \begin{tabular}{c|c|c}
    \toprule
    Method & PSPNet-50 &  Ours \\
    \midrule
    mIoU &43.65   & \textbf{46.42} \\
    \bottomrule
  \end{tabular}
}
\vspace*{0.05in}
\caption{Semantic segmentation results on S3DIS.}
\label{tab:s3dis}
\vspace*{-3.5mm}
\end{table}

\vspace*{-3.5mm}
\paragraph{S3DIS semantic segmentation.}~\label{sec:s3dis} 
Further, we experiment on S3IDS~\cite{armeni2017joint} with PointWeb~\cite{zhao2019pointweb} as the 3D network for extracting 3D features.
Results in Table~\ref{tab:s3dis} show that our method helps improve the baseline model, suggesting the generality and effectiveness of our method for use with different 3D network architectures on a different dataset.



\begin{figure*}[t]
\centering
\includegraphics[width=0.95\textwidth]{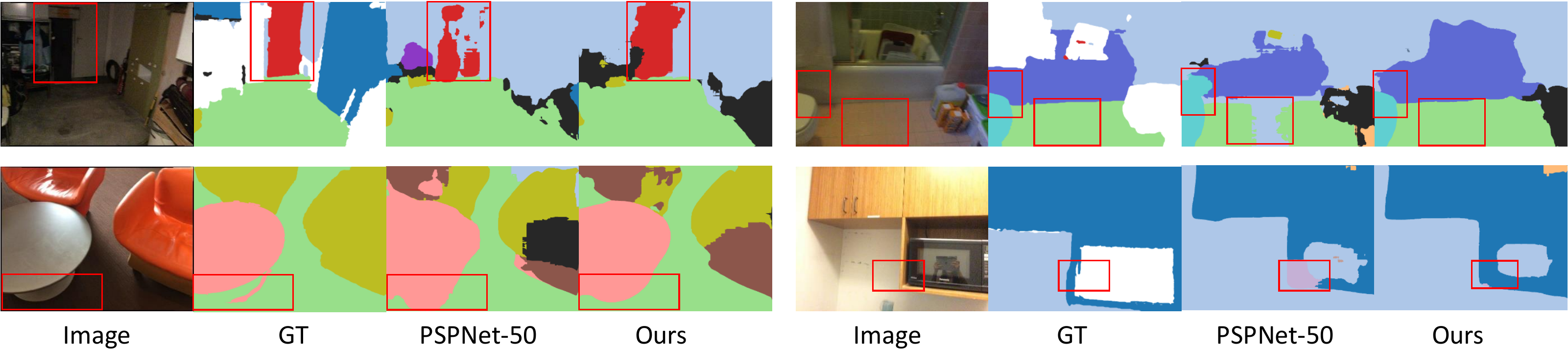}{}
\caption{Visualization of the results on the ScanNet-v2 dataset. As marked by the red boxes, our 3D-to-2D distillation has better performance for the door, floor, table, and cabinet regions benefited from the rich embedded 3D information.}
\label{fig:scannet}
\vspace*{-1mm}
\end{figure*}

\begin{figure*}[t]
\centering
\includegraphics[width=0.95\textwidth]{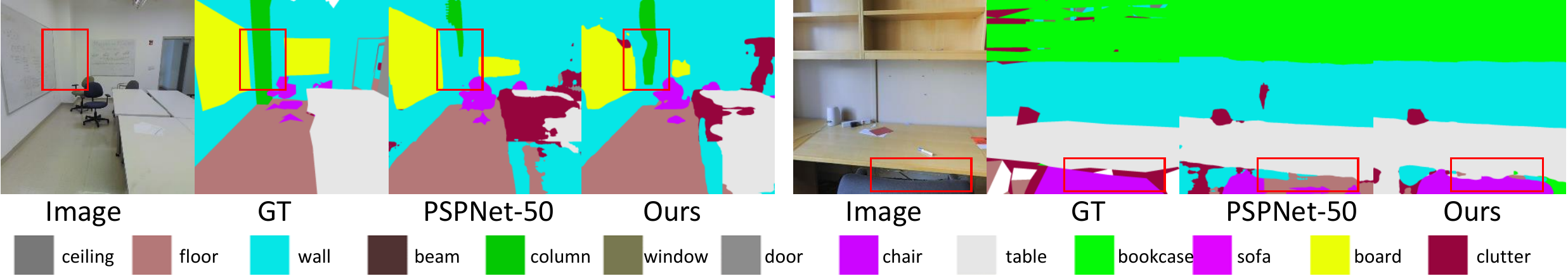}
\caption{Visualization of the results on S3DIS. Ours show the result of our approach using the ResNet-50 baseline as the backbone.}
\label{fig:s3dis}
\vspace*{-2.5mm}
\end{figure*}

\vspace*{-3.5mm}
\paragraph {Qualitative analysis.} 
Visual results on ScanNet-v2 and S3DIS are shown in Figures~\ref{fig:scannet} and~\ref{fig:s3dis}, respectively. The semantic segmentation quality has been consistently improved with our approach. The failure modes of the original PSPNet model are largely caused by occlusions (Figure~\ref{fig:scannet}: door, bathtub and toilet, Figure~\ref{fig:s3dis}: chair and table), uncommon viewpoints (Figure~\ref{fig:scannet}: table), and confusions due to similar color and texture (Figure~\ref{fig:s3dis}: column, and chair). In comparison, our approach can successfully segment these objects. 
This demonstrates the potential of our distilled 3D feature for resolving issues with occlusions, viewpoints, and textures.

This demonstrates that our model with distilled 3D feature can potentially  better leverage 3D geometric information, such as the shape of objects and thus is more robust to ambiguities caused by occlusions, viewpoints and texture. 

\begin{figure*}[t]
\centering
\includegraphics[width=0.99\textwidth]{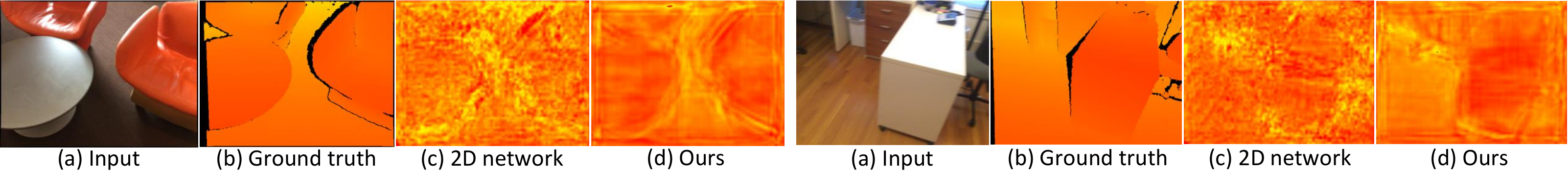}
\caption{Visualization of depth reconstruction quality. 
Depth map of PSPNet-50 and ours are both derived from  ``res-block-4'' features. }
\label{fig:depth2}
\vspace*{-2.5mm}
\end{figure*}

\begin{table}
\centering
\scalebox{0.8}{
  \begin{tabular}{c|cccc}
    \toprule
    Method & PSPNet-50  & Ours  \\
    \midrule
    mIoU & 49.50 & \textbf{51.70} \\
    \bottomrule
  \end{tabular}
}
\vspace{0.1in}
\caption{Results on NYU-v2 20 class without paired data.} 
\label{tab:unpair}
\vspace*{-2.5mm}
\end{table}


\subsection{3D-to-2D Distillation without Paired Data}

When paired data is not available, we generally can only train the 2D network solely with 2D image inputs.
With our adversarial training model (Section~\ref{sec:adverse}), we may distill unpaired 3D data to enrich features in the 2D network.
This subsection presents an experiment to compare the performance of a 2D network (\ie, PSPNet~\cite{zhao2017pyramid}) when it is trained (i) without 3D data and (ii) with unpaired 3D data.
Here, we use MinkowskiNet~\cite{choy20194d} as the 3D network, NYU-v2 as the 2D data, ScanNet-v2 as the 3D data, and the 20 categories common to both data in training and testing.

From Table~\ref{tab:unpair}, we can see that without using any 3D data, the average performance of PSPNet on NYU-v2 20 classes is 49.50, and using unpaired 3D data (with almost negligible effort) can enrich the 2D features and improve the performance by almost 5\% relatively,~\ie, from 49.50 to 51.70.
Note that this is the very first work that explores the potential of using unpaired 3D data for 2D semantic segmentation.
We did not explore more sophisticated techniques to further improve the results, but still, the results demonstrate the possibility of unpaired data, and we hope that this can open up a new direction for improving scene parsing from images.




\begin{table}
\centering
\scalebox{0.8}{
  \begin{tabular}{c|cc}
    \toprule
    Dataset & Ours (w/ BN) & Ours (w/ DN)    \\
    \midrule
     ScanNet-v2 &  57.64 & \textbf{58.22} \\
     S3DIS  &  45.57& \textbf{46.42} \\
      NYU-V2$^\dagger$ &  26.54 & \textbf{27.22} \\
    \midrule
     Dataset & Ours (w/o semantic) & Ours (w/ SAAL)    \\
    \midrule
    NYU-v2$^*$ &  50.86 & \textbf{51.84} \\
    \bottomrule
  \end{tabular}
}
\vspace{0.05in}
\caption{Ablation studies. NYU-v2$^\dagger$ is evaluated under the setting that the model is trained on ScanNet-V2 and tested on NYU-v2.
NYU-v2$^*$ is evaluated with unpaired 2D-3D data.
ScanNet-v2 and S3IDS results are evaluated on the setting with paired 2D-3D data.
SAAL denotes our semantic aware adversarial loss.}
\label{tab:ablationstudy}
\vspace*{-2.5mm}
\end{table}

\begin{figure}
\centering
\includegraphics[width=0.47\textwidth]{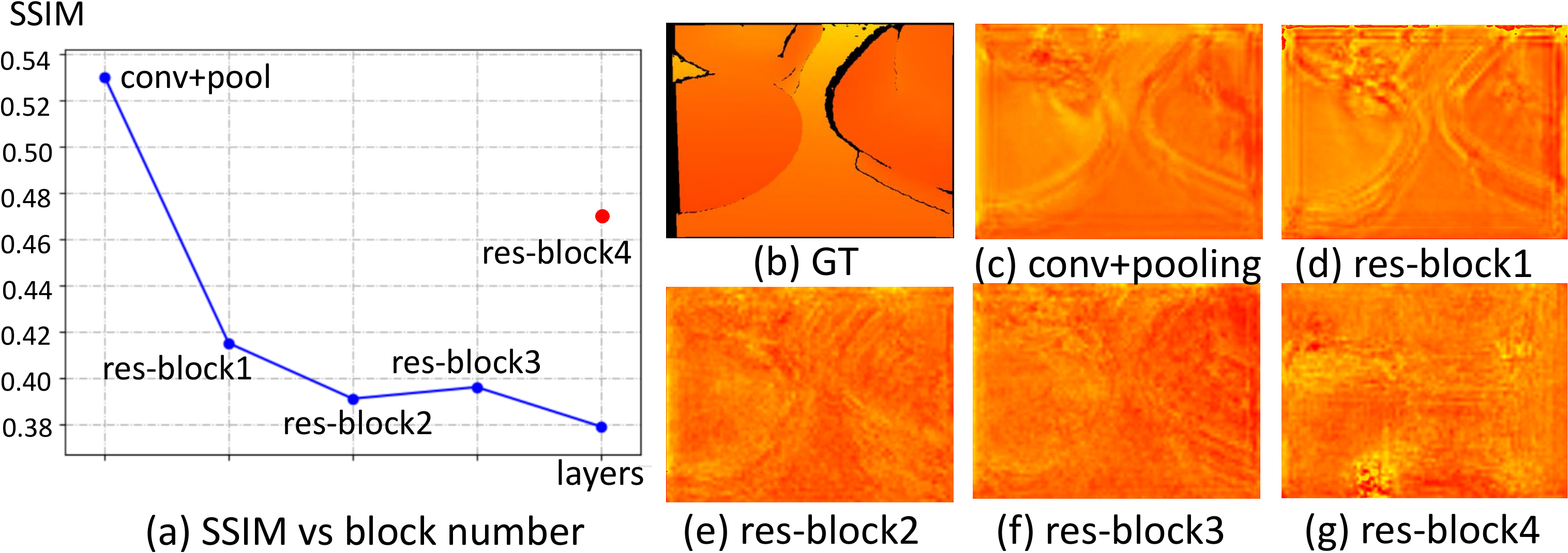}
\caption{PSPNet-50 retains less 3D information in deep layers {\vs} shallow layers.
(a) SSIM (larger is better) of depth images predicted from PSPNet-50 feature map in different layers.
(b)-(g): Depth maps predicted from different layers.
The PSPNet-50 is pre-trained for semantic segmentation.
Please zoom in to see the details.} \label{fig:depth1}
\vspace*{-2.5mm}
\end{figure}

\vspace{-0.35in}

\subsection{Ablation Studies} \label{sec:ablation}
Next, we ablate two major components in our approach:
(i) the dimension normalization modules---we replace the two DN modules with BNs while keeping the convolution part in DNs;
(ii) the semantic aware adversarial loss (SAAL)---we create a baseline that adopts a shared discriminator for all the categories, similar to that in~\cite{liu2020structured}. 

From the results shown in Table~\ref{tab:ablationstudy}, by comparing ``Ours (w/ BN)'' {\vs} ``Ours (w/ DN)'', we can see that DN consistently outperforms BN including the generalization to unseen domains,~\ie, NYU-v2$^\dagger$.
Then, by comparing our SAAL with the ablated semantic aware design in the NYU-v2$^*$ row, we can see that SAAL boosts the performance, implying that our adversarial training model helps improve the discriminative ability of the features, while aligning the distributions to facilitate feature transfer.

%

\vspace{-0.1in}

\subsection{Analysis}
\label{sec:motivation}
Further, we conduct an analysis to investigate whether our 3D-to-2D distillation can embed 3D information into the 2D CNN features and whether the 3D features can improve the model's generalizability. To start, we propose a metric to evaluate how much 3D information is embedded into the 2D CNN features.
%
Our idea is to evaluate the ability of CNN features for reconstructing depth maps. 
We build a depth estimation network with features from each layer as input and train it to produce the corresponding depth map. Then, we evaluate the depth reconstruction quality with the Structural Similarity Index (SSIM)~\cite{wang2004image}---larger is better. The reason for adopting SSIM instead of Mean Square Error is that depth information from a single image has scale ambiguities, while SSIM focuses more on structural similarities with the ground truth rather than the absolute values.

\textbf{3D features in baseline PSPNet}.
Taking inputs from different layers of PSPNet-50 trained on ScanNet-v2 for semantic segmentation, we obtain the depth prediction results shown in Figure~\ref{fig:depth1}.
Figure~\ref{fig:depth1} (a) shows that the SSIM value decreases as the feature map goes deeper. The associated visualizations in Figures~\ref{fig:depth1} (b)-(g) show that the structure of the depth map becomes hard to identify when it is derived from deep features (Figure~\ref{fig:depth1} (g)).
Both quantitative and qualitative results suggest that deep features may retain less 3D information and have a lower capability of reconstructing depth in comparison with shallow features, even though they are more directly relevant to the final semantic parsing task.

\textbf{3D features in our models.} Further, we analyze whether the 2D network in our framework can better leverage 3D information. We take the deep ``res-block-4'' feature map of our model trained on ScanNet-v2. The SSIM is improved by $24\%$ (baseline: 0.38 {\vs} our: 0.47) as compared with the baseline (the red point in Figure~\ref{fig:depth1}), and the quality of the reconstructed depth map is significantly improved as shown in Figure~\ref{fig:depth2}. The object structure and boundary can be better preserved. With 3D-to-2D distillation, higher quality depth can be reconstructed from our deep features, suggesting that our approach facilitates the 2D network to better utilize 3D information when constructing the deep features.

\begin{table}
\centering
\scalebox{0.8}{
  \begin{tabular}{c|ccccc}
    \toprule
    Method & PSPNet-50 & PSPNet-101  & Multitask & Cascade & Ours  \\
    \midrule
    mIoU & 19.08 &  20.12   &20.30& 22.52 & \textbf{27.22} \\
    \bottomrule
  \end{tabular}
}
\vspace*{0.01in}
\caption{Domain generalization results on NYU-v2 20-class using various models trained on ScanNet-v2.}
\label{tab:domain}
\end{table}
\subsection{Effectiveness in Boosting Model Generalization}


The analysis in Section~\ref{sec:motivation} implies that the embedded 3D information in deep features may help the CNN better utilize the robust 3D cues, such as shape, for recognition, and further improve the model's generalization abilities. 
Next, we investigate whether 3D-to-2D distillation can improve the generalizability of the model. To this end, 
we directly evaluate the baseline model and our model when trained on ScanNet-v2 and tested on NYU-v2.
Here, we consider only the 20 classes common to ScanNet-v2 and NYU-v2.

Comparing Tables~\ref{tab:domain} and~\ref{tab:scannet},
we can see that the performance of all methods drops seriously when tested on unseen NYU-v2. 
However, with the 3D-enhanced 2D features, our model with the ResNet-50 backbone improves over PSPNet-50 by 43\% for more than 8\% mIoU.
Also, it surpasses PSPNet-101 by more than 7\% mIoU.
The results imply that the embedded 3D feature helps improve the generalizability of the 2D network, and such improvement cannot be achieved by using a larger network,~\ie, PSPNet-101.

\section{Conclusion}
This paper presents a novel 3D-to-2D distillation framework that effectively leverages 3D features learned from 3D point clouds to enhance 2D networks for indoor scene parsing. At testing, the 2D network can infer simulated 3D features without any 3D data input. To bridge the statistical distribution gap between the 2D and 3D features, we propose a two-stage distillation normalization module for effective feature integration. Further, to broaden the applicability of our approach, we design an adversarial training model with the semantic aware adversarial loss to extend our framework for training with unpaired 2D-3D data.
Experiments on three public indoor datasets suggest the superiority of our approach in various settings.
We hope our further analysis on 3D and generalization could inspire future works on incorporating 3D information to improve model generalization.

\clearpage
\appendix
\centerline{\Large{\textbf{Supplementary Material}}}

In this supplementary material, we show sparse 3D data projected onto associated 2D images (Section~\ref{sec:proj}), 
quantitative (per-category IoU) and qualitative comparisons for unpaired 3D-to-2D distillation (Section~\ref{sec:unpair}),
visual results on NYU-v2 to show the generalizability of our method (Section~\ref{sec:unseen}),
the configurations in the depth prediction task (Section~\ref{sec:depth}), 
and the network architecture of the discriminator in Semantic Aware Adversarial Loss (Section~\ref{sec:SAAL}).
The code can be found in \url{https://github.com/liuzhengzhe/3D-to-2D-Distillation-for-Indoor-Scene-Parsing}. 



\section{Visualization of the Projected 3D data to 2D Image}\label{sec:proj}
This section corresponds to Section 4.2 in the main paper.
In this section, we show 3D data projected onto associated 2D images.

We use full 3D data for feature extraction, but after projecting all points to the 2D image domain, we save only the 3D point features in the 2D image grid at $(x,y)$ pixel locations that are multiples of $K$ (empirically set as 8) to reduce the I/O burden when training the 2D network.
This is certainly a trade-off between the I/O burden and the performance.

\begin{figure*}[!t]
\centering
\includegraphics[width=1\textwidth]{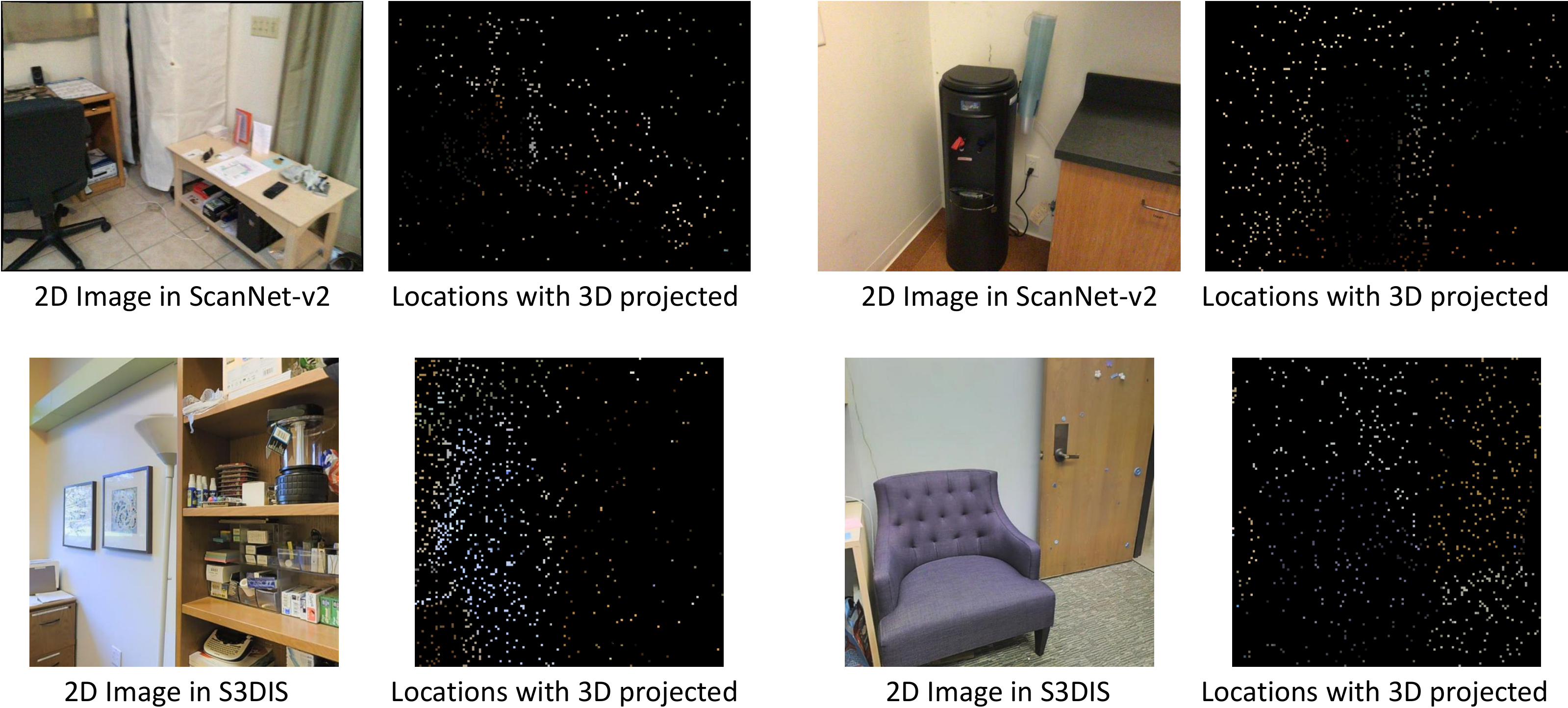}
\caption{Projecting the sparse 3D data onto the associated 2D images for the sparse paired 2D-3D data in ScanNet-v2 (top row) and S3DIS (bottom row).
We only perform the regression as Equation 1 on the pixels with projected 3D data.}
\label{fig:project}
\end{figure*}

As shown in Figure~\ref{fig:project}, we can see that the projected 3D data points are very sparse, even though they are dilated to increase the number of pixels occupied by each 3D point.
On average, only $16.38\%$ and $10.57\%$ of 2D pixels have corresponding 3D points in ScanNet-v2 and in S3DIS, respectively. 

\section{Detailed Results and Visualization of the Unpaired 3D-2D distillation results }\label{sec:unpair}

\begin{figure*}[!t]
\centering
\includegraphics[width=1\textwidth]{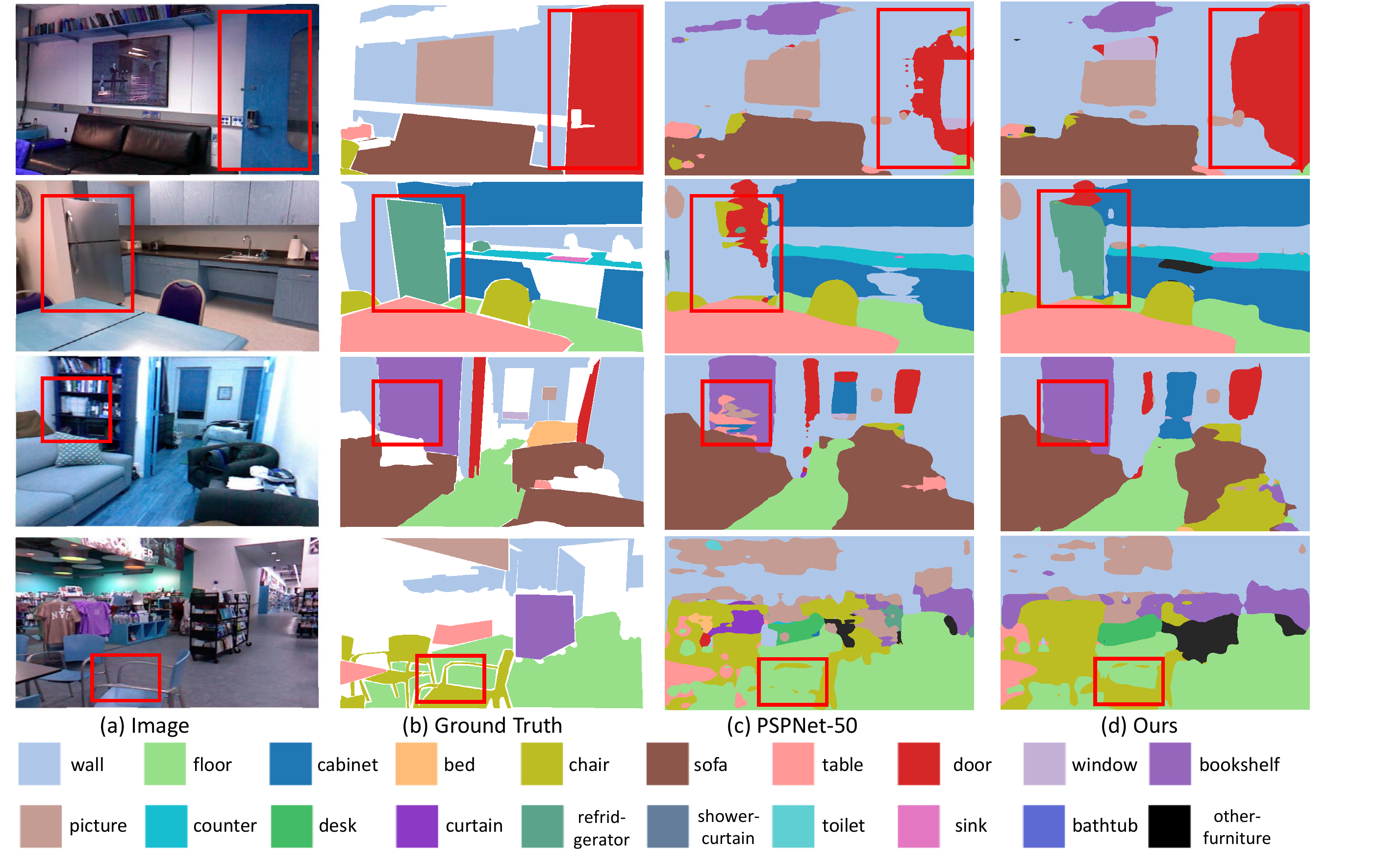}
\caption{Visual comparison on the NYU-v2 data, with and without using our 3D-to-2D distillation method. Comparing the results shown on the rightmost two columns, we can see that our 3D-to-2D distillation model better predicts the door, the refrigerator, the bookshelf, and the chair.
Our model makes these predictions by leveraging the 3D information distilled from the ``ScanNet-v2'' 3D scenes, which are {\em not paired\/} with the NYU-v2 image data shown above.}
\label{fig:unpair}
\end{figure*}

This section corresponds to Section 4.3 in the paper. Here, we present the per-category performance and quantitative results of our 3D-to-2D distillation on unpaired 2D-3D data. As shown in Table~\ref{tab:unpair}, the results of most categories can be improved using our method.
Further, as Figure~\ref{fig:unpair} shows, PSPNet (baseline) has some failure cases, especially when the 2D image cues are confusing or misleading. For example, as marked by the red boxes in Figure~\ref{fig:unpair}, the door in row 1 has an unusual blue color and the refrigerator in row 2 has a similar color and texture as the door. The bookshelf in row 3 is partially occluded and the chair in row 4 looks similar to the floor. In comparison, our 3D-to-2D distillation model improves PSPNet (baseline).
%
This demonstrates that our approach enhances the 2D network to better leverage the geometrical information for resolving issues like occlusions, viewpoints, and texture, even without paired 2D-3D data.


\begin{table*}[!t]
\centering
\scalebox{0.7}{
  \begin{tabular}{c|c|ccccccccccc}
    \toprule
    \multirow{2}{*}{Method}  &	\multirow{2}{*}{mIoU}	& wall & floor &	cabinet&	bed	&chair&	sofa&	table&	door&	window&	bookshelf \\
    &&picture& counter&	desk&	curtain& refrigerator&	shower curtain&	toilet&	sink&	bathtub&	other-furniture\\
    \midrule
    \multirow{2}{*}{PSPNet-50~\cite{zhao2017pyramid}}  & \multirow{2}{*}{49.50} & 79.06&	84.81& 61.62&	69.73&	60.13&	60.53&	41.10&	36.28&	54.52&	54.53\\	
    && 62.70&	57.30&	15.37&	48.74	&36.86&	14.25&	62.65&	49.35&24.59&	16.03&	\\
	\midrule
	PSPNet-50  &  \multirow{2}{*}{\textbf{51.70}}  &	80.11 &	83.28 & 60.96 &	71.57 &	61.19 &	 63.82 &	40.16 & 39.29 &	55.48 & 56.22\\
	+ \textbf{Our Unpaired} 3D-to-2D Distillation && 64.28 & 58.80 & 17.63 & 49.12&	45.84& 22.43&	67.67&	53.38&	25.00&	17.61	\\
    \bottomrule
  \end{tabular}
}
\vspace*{2mm}
\caption{Unpaired 3D-to-2D distillation results on the ScanNet-v2 validation set.
In the 2nd column from the left, we compare the overall performance (mIoU) of the baseline (PSPNet-50) vs. our full model (bottom), and then in the subsequent ten columns, we compare the per-category performance (20 categories in total), where in each of these columns, we show results for two categories in each cell.
With almost negligible extra effort, our approach can improve the results of most of the categories, and relatively improve the mean IoU by almost 5\%,~\ie, from 49.50 to 51.70, by means of training the network with unpaired 2D-3D data.}
\label{tab:unpair}
\end{table*}

\section{Visualization of the Model Generalization Results on NYU-v2}\label{sec:unseen}
\begin{figure*}[hb]
\centering
\includegraphics[width=1\textwidth]{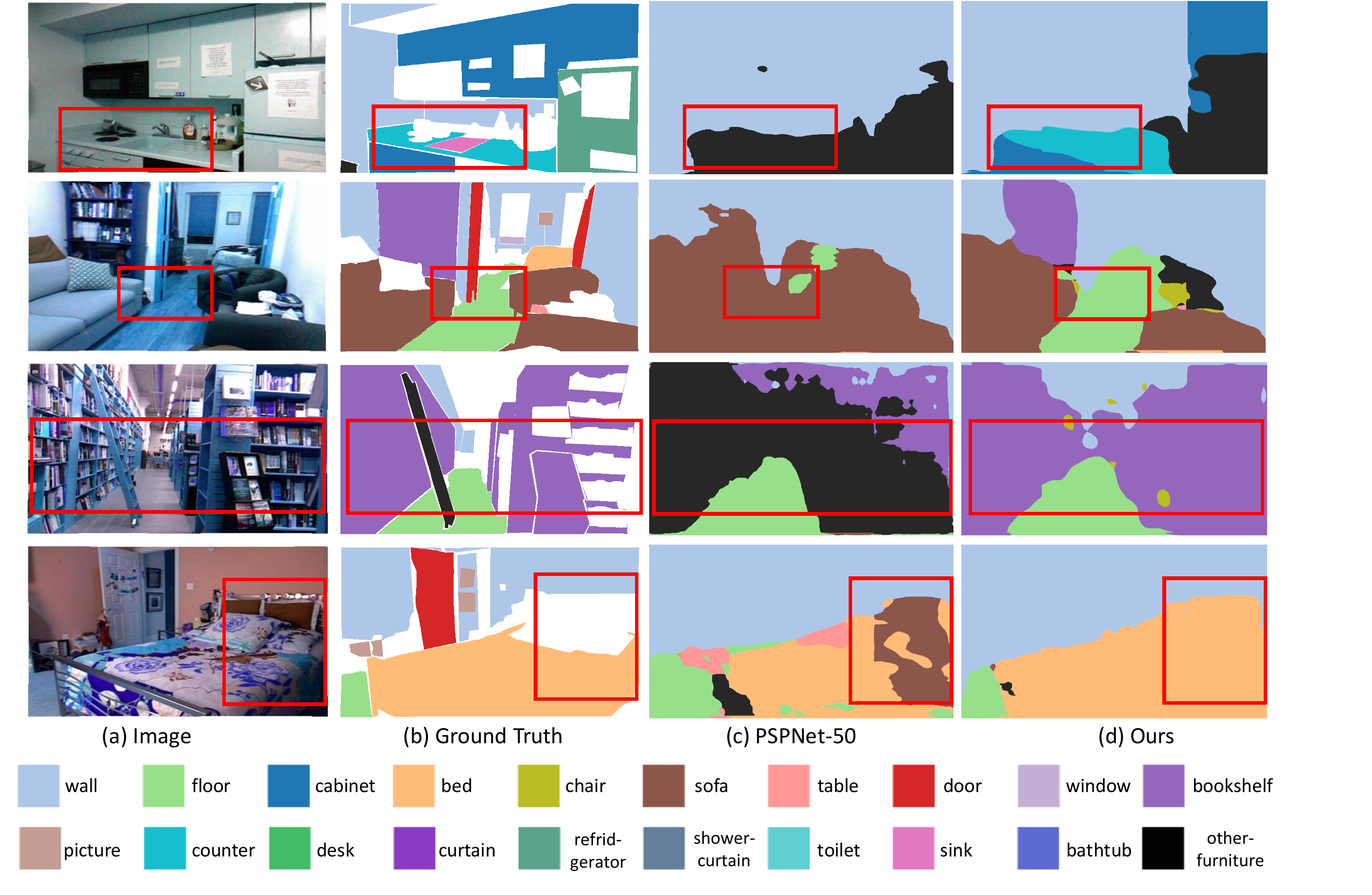}
\caption{Visualization of the results that demonstrate the generalizability of our model.
Both models,~\ie, (c) \& (d), are trained on ScanNet-v2 and tested on NYU-v2. Our model (d) with the 3D-to-2D distillation can better predict the cabinet, counter, bookshelf, floor, and bed (see the red boxes) by leveraging the distilled  3D information.
Quantitative comparison results can be found in the main paper.} \label{fig:unseen}
\end{figure*}

This section corresponds to Section 4.6 in the main paper.
Here, we show the visual results of the baseline and our model, both trained only on ScanNet-v2 and directly tested on NYU-v2. The results in Figure~\ref{fig:unseen}(d) manifest that our model can better segment the objects in the unseen NYU-v2 dataset by leveraging the distilled 3D information.
The results demonstrate the generalization ability of our 3D-to-2D distillation model.

\section{Details of Depth Prediction for 3D Information Evaluation}\label{sec:depth}

This section provides details for Section 4.5 in the main paper. 
In detail, the depth prediction network uses VGG~\cite{simonyan2014very} as the backbone.
We use the first 2k images in the ScanNet-v2 training set for training and the first 200 images in the ScanNet-v2 validation set for testing. All models (including the baseline) are trained with SGD for 10 epochs. The initial learning rate is $5e^{-7}$ and decreased by $2.5e^{-8}$ for each epoch. 
Figure~\ref{fig:depth2} shows two more examples that follow the style of Figure 7 in the main paper, showing that the depth information reconstructed from our network has better 3D structure than the depth information reconstructed from the baseline 2D network without our 3D-to-2D distillation.

\begin{figure*}[hb]
\centering
\includegraphics[width=1\textwidth]{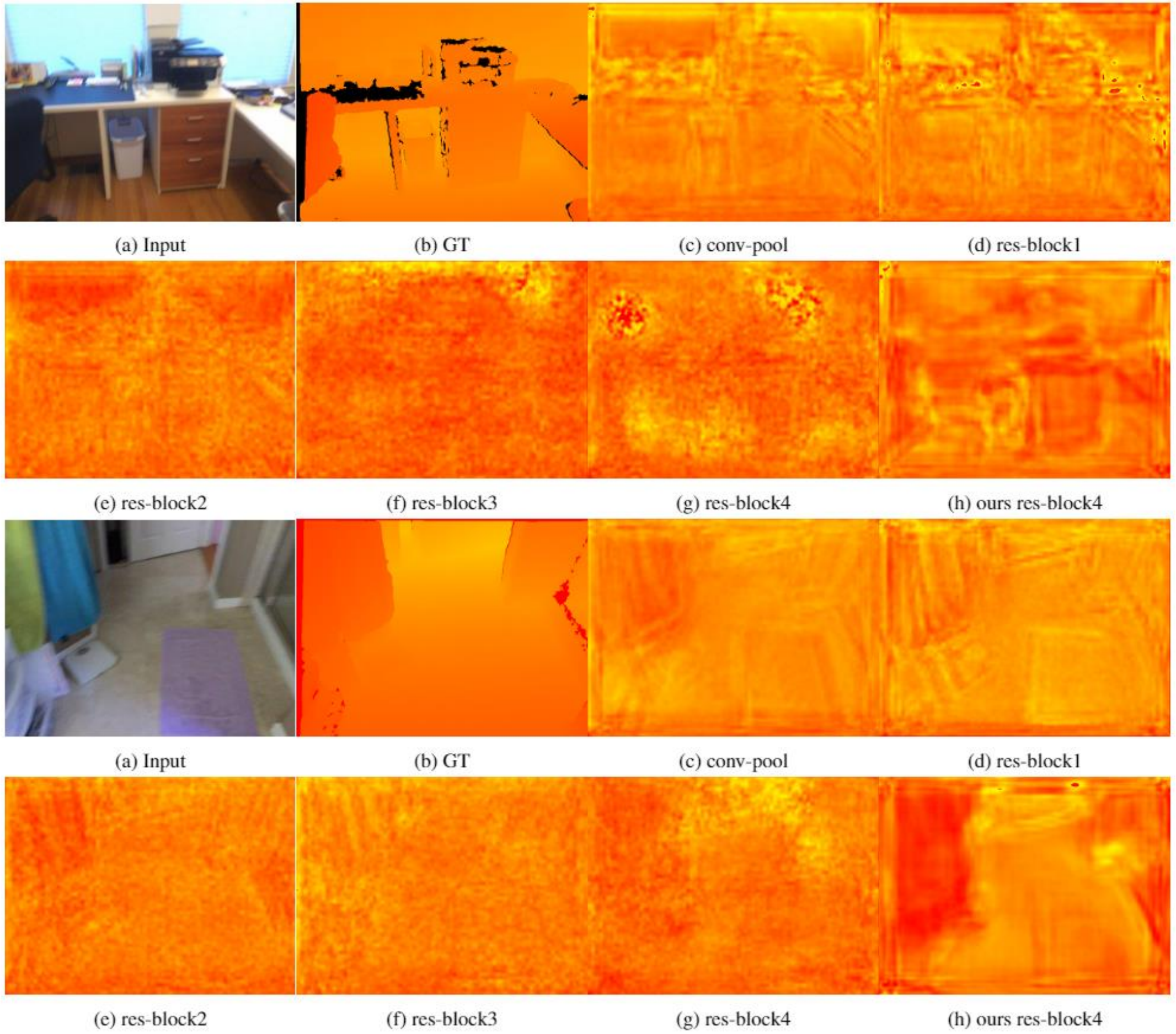}
\caption{Depth reconstructed from the original 2D network as baseline (c-g) and ours (h) vs. the ground truth (GT) shown in (b).
(c-g) are from the feature maps of PSPNet-50, whereas (h) is from the ``res-block-4'' of the network with our 3D-to-2D distillation.} \label{fig:depth2}
\end{figure*}


\section{Network Architecture of the Discriminator of Semantic-Aware Adversarial Loss}
\label{sec:SAAL}

\vspace{-0.12in}
The architecture of discriminator $D^c$ is composed of $6$ fully-connected layers with channels 64, 32, 16, 8, 4, and 1, respectively. Each of the first five is followed by a Leaky-ReLU~\cite{xu2015empirical} as the activation function, and the last one is followed by a Sigmoid operation to modulate the output within the interval $(0,1)$, indicating the confidence of whether the input feature vector is from the 2D network or 3D network.
$D^c$ is trained with all the feature vectors belonging to category $c$ and the discriminators for different categories are optimized individually without sharing weights. 
Each discriminator only has 0.009M parameter, and its input is an 8$\times$ down-sampled feature map with $96$ channels, the computation is 63.9M MAC for all of them, and all discriminators are trained simultaneously.

{\small
\bibliographystyle{ieee_fullname}
\bibliography{cvpr}
}

\end{document}